# Examining the Classification Accuracy of TSVMs with Feature Selection in Comparison with the GLAD Algorithm

Hala Helmi, Jonathan M. Garibaldi and Uwe Aickelin

*Abstract-* Gene expression data sets are used to classify and predict patient diagnostic categories. As we know, it is extremely difficult and expensive to obtain gene expression labelled examples. Moreover, conventional supervised approaches cannot function properly when labelled data (training examples) are insufficient using Support Vector Machines (SVM) algorithms. Therefore, in this paper, we suggest Transductive Support Vector Machines (TSVMs) as semi-supervised learning algorithms, learning with both labelled samples data and unlabelled samples to perform the classification of microarray data. To prune the superfluous genes and samples we used a feature selection method called Recursive Feature Elimination (RFE), which is supposed to enhance the output of classification and avoid the local optimization problem. We examined the classification prediction accuracy of the TSVM-RFE algorithm in comparison with the Genetic Learning Across Datasets (GLAD) algorithm, as both are semi-supervised learning methods. Comparing these two methods, we found that the TSVM-RFE surpassed both a SVM using RFE and GLAD.

## I. INTRODUCTION

Data mining techniques have traditionally been used to extract hidden predictive information in many diverse contexts. Usually datasets contain thousands of examples. Recently the growth in biology, medical science, and DNA analysis has led to the accumulation of vast amounts of biomedical data that require in-depth analysis.

After years of research and development, many data mining, machine learning, statistical analysis systems and tools are available to be used in biodata analysis. Consequently, this paper will examine a relatively new technique in data mining. This technique is called Transductive Supervised Support Vector Machines [2], also named Semi-Supervised Support Vector Machines $S^3$VMs, located between supervised learning with fully-labelled training data and unsupervised learning without any labelled training data [1]. In this method, we used both labelled and unlabelled samples for training: a small amount of labelled data and a large amount of unlabelled data.

Hala Helmi (PhD student), Jon Garibaldi (Assoc. Prof.) and Professor Uwe Aickelin are with the Intelligent Modelling and Analysis (IMA) Research Group, School of Computer Science, The University of Nottingham, Jubilee Campus, Wollaton Road, Nottingham, NG8 1BB, U.K.
{hqh, jmg, uxa} @cs.nott.ac.uk

The purpose of this paper is to observe the performance of Transductive SVMs combined with a feature selection method called Recursive Feature Elimination (RFE), which is used to select molecular descriptors for Transductive Support Vector Machines (TSVMs).

The paper is organized as follows: section 2 provides a literature review on Support Vector Machines, Transductive Support Vector Machines and finally Recursive Feature Elimination. In section 3, the TSVM algorithm combined with RFE is detailed, as well as a brief summary of the GLAD algorithm based on a recently published paper [3] which aims to compare the prediction accuracy of these two algorithms. Section 4 is dedicated to comparing and analysing the experimental results of both algorithms (TSVM and GLAD). Finally, a summary of the results and discussion will be presented in section 5.

## II. BACKGROUND

### A. Support Vector Machines

Support Vector Machine (SVMs), as a supervised machine learning technique, perform well in several areas of biological research, including evaluating microarray expression data [4], detecting remote protein homologies [5] and recognizing translation initiation sites [6]. SVMs have demonstrated the ability not only to separate the entities correctly into appropriate classes, but also to identify instances where established classification is not supported by the data [7]. SVMs are a technique that makes use of training that utilizes samples to determine beforehand which data should be clustered together [4].

### B. Tranductive Support Vector Machines

Transductive learning is a method strongly connected to semi-supervised learning, where semi-supervised learning is intermediate between supervised and unsupervised learning. Vapnik introduced Semi-Supervised Learning for Support Vector Machines in the 1990s. His view was that transduction (TSVM) is preferable to induction (SVM), since induction needs to solve a more general problem (inferring a function) before solving a more detailed one (computing outputs for new cases) [8] [9].

Transductive Support Vector Machines attempt to maximize the hyperplane classifier between two classes using labelled training data; at the same time this forces the hyperplane to be far away from the unlabelled samples. TSVMs seem to be a perfect semi-supervised learning algorithm because they combine the regularization of Support Vector Machines with a straight forward implementation of the clustering assumption [10].

## C. Recursive Feature Elimination

Most prediction model algorithms are less effective when the size of the data set is large. There are several methods for decreasing the amount of the feature set. From among these methods we selected a technique called Recursive Feature Elimination (RFE). The basis for RFE is to begin with all the features, select the least useful, remove this feature, and then repeat until some stopping condition is reached.

Finding the best subset features is too expensive, so RFE decreases the difficulty of feature selection by being 'greedy' [11].

## III. METHODS

This section of the paper is focused on describing TSVM-RFE, the problem motivated by the task of classifying biomedical data. The goal is to examine classifier accuracy and classification errors using the Transductive Support Vector Machines method, in order to determine whether this method is an effective model when combined with Recursive Feature Elimination (RFE) compared with the Genetic Learning Across Datasets (GLAD) algorithm.

### A. Support Vector Machines

The purpose of SVMs is to locate a classifier with the greatest margin between the samples relating to two different classes, where the training error is minimized. Therefore, to achieve this we used a set of $n$-dimensional training samples $X = \{x_i\}_{i=1}^{m}$ labelled $\{y_i\}_{i=1}^{m}$ and their mapping $\{\ (x_i)\}_{i=1}^{m}$ via kernel function:

$$k(x_i, x_j) = (x_i)^T (x_j)'$$

SVM has the following primal form:

Minimize over $(w, b, \xi_1, \ldots, \xi_m)$

$$\|w\|_p^p + C \sum_{i=1}^{m} \xi_i$$

Subject to:
$$\forall_{i=1}^{m}: y_i (w^T (z * x_i)' + b) \geq 1 - \xi_i, \xi_i \geq 0 \quad (1)$$

The SVM predictor for samples x, as shown below, was settled on by the vector inner product between the **w** and the mapped vector (x), plus the constant $b$.

$$y = \text{sgn}(w^T (x)' + b)$$

The predictor actually corresponds to a separating hyperplane in the mapped feature space. The prediction for each training samples $x_i$ is connected with a violation term $\xi_i$. The C is a user-specified constant to manage the penalty for these violation terms.

The parameter p in the above (1) points to which kind of norm of **w** is assessed. It is usually set to 1 or 2, resulting in the 1-norm ($l_1$-SVM) and 2-norm SVM ($l_2$-SVM) respectively. The 1-norm and 2-norm TSVMs have been discussed in [12] and [13].

### B. Transductive Support Vector Machines

In this paper, we are using the extended SVM technique of transductive SVMs and we methodically adept the 2-norm for the TSVM.

The standard setting can be illustrated as:

Minimize over $(y_1^*, \ldots, y_k^*, w, b, \xi_1, \ldots, \xi_m, \xi_1^*)$

$$\frac{1}{2}\|w\|_2^2 + C \sum_{i=1}^{m} \xi_i + C^* \sum_{j=1}^{k} \xi_j^*$$

subject to:
$$\forall_{i=1}^{m}: y_i (w^T (z * x_i)' + b) \geq 1 - \xi_i, \xi_i \geq 0$$
$$\forall_{j=1}^{k}: y_j^* (w^T (z * x_j^*)' + b) \geq 1 - \xi_j^*, \xi_j^* \geq 0 \quad (2)$$

Where each $y_j^*$ is the unknown label for $x_j^*$ which is one of the k unlabelled samples; compared with SVM (1), the formulation (2) of the TSVMs takes the unlabelled data into consideration by representing the violation terms $\xi_j^*$ caused by forecasting each unlabelled pattern $(x_j^*)$ into $y_j^*$. The penalty for these violation terms is controlled by a new constant $C^*$ labelled with unlabelled samples, while C consists of labelled samples only.

Precisely solving the transductive problem needs a search of all potential assignments of $y_1^*, \ldots, y_k^*$ and identifying the various terms of $\xi^*$ which are regularly intractable for large data sets. It is worth mentioning the $l_2$-TSVM implemented in the SVMLight [18] [8].

### C. Recursive Feature Elimination

Recursive Feature Elimination (RFE) has the advantage of decreasing the number of redundant and recursive features. RFE decreases the difficulty of feature selection by being greedy.

To extend SVM feature selection techniques to transductive feature selection is specifically straightforward, as we can produce TSVM-RFE by iteratively eliminating features with weights calculated from TSVM models. We can explain the TSVM-RFE approach as the following standard process.

1. Pre-process data and calculate filtering scores $\bar{s}$. Moreover, optionally further normalize data. This approach first filters some features based on scores like Pearson correlation coefficients.
2. Adjust **z** as an all-one input vector.
3. Set $z \leftarrow z * \bar{s}$. Set part of the small entries of **z** zero according to a proportion/threshold, and probably discrete non-zero **z** to 1.
4. Obtain a (sub-) optimal TSVM as calculated by cross-validation accuracy (2).
5. For RFE approaches, estimate feature weights $\bar{s}$ from the model in step 4 according to:

$$J_t' = -\frac{1}{2} \sum_{i,j}^{m} \alpha_i \alpha_j y_i y_j (x_i / x_{it})^T (x_j, x_{jt})' + \sum_{i=1}^{n} \alpha_i \quad (3)$$

Where $(x_i/x_{it})$ indicates the input samples $i$ with feature $t$ removed. The weight of the $t$-th feature can be clarified as

$$s_t = \sqrt{|\Delta J_t|} = \sqrt{|J - J'_t|}$$

The following estimation suggested in [11] is easier to measure.

$$s_t^2 \approx \sum_{i,j=1}^{m} \alpha_i \alpha_j \mathcal{Y}_i \mathcal{Y}_j \ (x_{it})^T (x_{jt})'$$

Specifically, the feature weights are identical to the **w** if the SVM is built upon a linear kernel.

Return to step 3 unless there is an acceptable number of features/iterations. Output the closing predictor and features highlighted by large values of z.

Step 3 comprises selecting a proportion/number of features according to a threshold cutting the vector z. For filtering scores and the RFE method, the vector z is changed to a binary vector. Then the $z * x$ has the effect of pruning or deactivating some features.

The threshold is usually found to prune a (fixed) number/proportion of features at each iteration. The value of the remaining features is then measured by the optimality of the TSVM model obtained in step 4. We then apply cross-validation accuracy as the performance measure for the TSVM algorithm. For a subset of features selected by choosing a threshold value, we extend the model search upon the free parameters, such as $[C, C^*\sigma (RBF), d(Poly)]$ and choose the preferable parameter set which results in the highest cross-validation accuracy.

### D. Genetic Learning Across Datasets (GLAD)

The GLAD algorithm is different from prior algorithms of semi-supervised learning. The GLAD algorithm has been applied as a wrapper method for feature selection. A Genetic Algorithm (GA) was implemented for generating a population of related feature subsets. The labelled data and the unlabelled data samples were computed separately. Linear Discriminant Analysis (LDA) and K-means (K = 2) for these two data forms of cluster algorithms were used [14]. A distinctive two-term scoring function resulted to independently score the labelled and unlabelled data samples. Generally, the score was calculated as a weighted average of the two terms as shown below.

$$score = w \times score_{labelled} + (1 - w) \times score_{unlabelled} \quad (4)$$

As the typical leave-one-out-cross-validation accuracy for the labelled training samples, they identified the labelled data samples score. The unlabelled data samples score consists of two terms: a cluster separation term and a steady ratio term.

$$score_{unlabelled} = \frac{\sum_{i \neq j} |C_i - C_j|}{\sum_{i \neq j} |C_i - C_j| + \sum_i \frac{1}{N_{C_i}} \sum_j |x_{ij} - C_i|} -$$

$$\sqrt{\frac{1}{n_c} \sum_i (\pi_i - \pi_{\exp_i})^2} \quad (5)$$

$C_i$ = centroid of cluster; $\pi_i$ = ratio of data in cluster i ; $\pi_{\exp_i}$ = expected ratio in cluster i ; $N_{C_i}$ = number of data samples in cluster i; $n_c$ = number of clusters.

## IV. EXPERIMENTS AND RESULTS

This section discusses the results of the experiments which were carried out in order to assess the effectiveness of the classification model accuracy proposed in the previous section.

### A. Datasets

- **Leukaemia (AML-ALL):** including 7129 probes, two variants of leukaemia are available: acute myeloblastic leukaemia (AML), 25 samples; and acute lymphoblastic leukaemia (ALL), 47 samples [15].
- **Lymphoma (DLBCL):** consisting of 7129 genes and 58 DLBCL samples. Diffuse large B-cell lymphoma (DLBCL) and 19 samples of follicular lymphoma (FL) [16].
- **Chronic Myeloid Leukaemia (CML):** contained 30 samples (18 severe emphysema, 12 mild or no emphysema) with 22,283 human gene probe sets [17].

### B. TSVM Recursive Feature Elimination (TSVM-RFE) Result

- **Leukaemia (AML-ALL):** the results for the Leukaemia ALL/AML dataset are summarized in Figure 1 in the diagram on the left. TSVM-RFE gave the smallest minimal error of 3.68%, and compassionately smaller errors compared with SVM-RFE: 3.97% for 30, 40, ..., 70 genes. Interestingly, in our experiments both methods gave the lowest error when 60 genes were used. This provided a reasonable suggestion for the number of relevant genes that should be used for the leukaemia data.
- **Lymphoma (DLBCL):** the results for the Lymphoma (DLBCL) dataset are summarized in Figure 1 in the middle diagram. TSVM-RFE gave the smallest minimal error of 3.89%, and quite firmly smaller errors compared to the SVM-RFE: 4.72% for 30, 40, ..., 70 genes. For TSVM, the methods gave the lowest error for 60 genes, while SVM methods gave the lowest error at 50 genes with 4.72% compared to 4.97% for 60 genes. This could give a sensible suggestion for the number of relevant genes that should be used for the lymphoma (DLBCL) data.

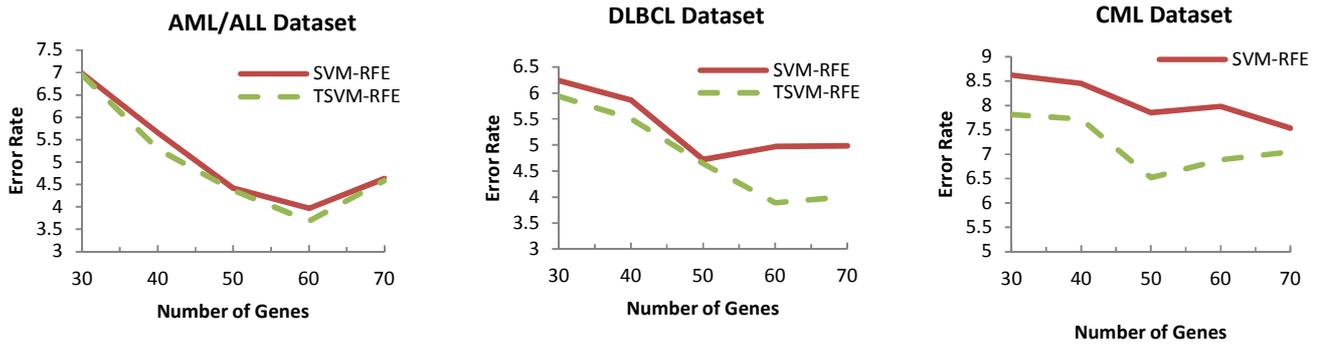

**Figure 1:** Testing error for three data sets. The 5-fold cross-validated pair t-test shows the SVM-RFE and the TSVM-RFE have relative differences when comparing the two methods at the confidence rate of 95% (Linear kernel, C = 1).

- **Leukaemia (CML):** lastly, the TSVM-RFE and SVM-RFE results for the Leukaemia (CML) dataset are provided in Figure 1 in the diagram on the right. TSVM-RFE gave the smallest minimal error of 6.52%, and critically smaller errors in contrast to the 7.85% SVM-RFE for 30, 40, ..., 70 genes. Both algorithms showed the lowest error when 50 genes were used. This presented a sensible proposal for the number of related genes that should be used for the Leukaemia (CML) data.

*C. Comparing the TSVM Algorithm result with the GLAD Algorithm*

Implementing Genetic Learning Across Datasets involved conducting three experiments using previous datasets, each addressing a different cancer diagnostic problem: ALL/AML for disparity in diagnosis; in CML a dataset predicting the response of imatinib; and in DLBCL for forecasting outcome.

In the AML-ALL dataset, the accuracy range using only labelled samples was 73.46%. Combining unlabelled samples with labelled samples increased the range to 75.14%. Adding unlabelled samples increased the accuracy from 59.34% to 65.57% in the CML experiments. The addition of the unlabelled samples to the unlabelled samples for DLBCL raised the accuracy from 49.67% to 55.79%. This shows that the GLAD algorithm outperformed the SVM-RFE and TSVM-RFE in some cases when we made use of the labelled data only without gene selection. In Table 1, for example the AML-ALL dataset, the GLAD algorithm gives 73.46% while SVM-RFE and TSVM-RFE accuracy are 52.8% and 55.6% respectively.

However, in the second dataset DLBCL showed that GLAD algorithm accuracy was 49.67% and SVM-RFE 55.8%. Furthermore, the third dataset of CML, SVM-RFE gave 59.02% without gene selection, while GLAD gave 59.34%. On the other hand, TSVM exceeded GLAD when making use of unlabelled data along with labelled data and selecting genes. The results are shown in Table 1.

For instance, for the CML dataset using all the samples without gene selection TSVM gave 72.6% when selecting genes based on REF, TSVM exceeded 93.48%, while GLAD gave 65.57% with gene selection. In the same vein, the accuracy for the DLBCL dataset achieved 96.11% by TSVM with gene selection.

On the other hand, the GLAD algorithm gave 55.79% with gene selection. As well as this, the TSVM with the AML-ALL dataset with gene selection gave 96.32% while the GLAD algorithm gave 75.14%. This means that the TSVM performed better than the GLAD algorithm, and the performance with gene selection showed a superior result.

**Table 1: Accuracy obtained with SVM-RFE, TSVM-RFE and GLAD**

| Dataset | | SVM-RFE Accuracy (labelled) | TSVM-RFE Accuracy | GLAD Accuracy |
|---|---|---|---|---|
| **ALL-AML** | | | | |
| Without Selection | 7219 Genes, 72 Samples | 52.8% | 55.6% | 73.46% (labelled) |
| With Selection | 60 Genes, 72 Samples | 96.03% | 96.32% | 75.14% |
| **DLBCL** | | | | |
| Without Selection | 7219 Genes, 77 Samples | 55.8% | 57.1% | 49.67% (labelled) |
| With Selection | 60 Genes, 77 Samples | 95.03% | 96.11% | 55.79% |
| **CML** | | | | |
| Without Selection | 22,283 Genes, 30 Samples | 59.02% | 72.6% | 59.34% (labelled) |
| With Selection | 50 Genes, 30 Samples | 92.15% | 93.48% | 65.57% |

V. CONCLUSION

This paper has investigated topics focused on semi-supervised learning. This was achieved by comparing two different methods for semi-supervised learning using previously classified cancer datasets.

The results on average for semi-supervised learning surpassed those for supervised learning. However, this shows that the GLAD algorithm outperformed SVM-RFE when we made use of the labelled data only. On the other hand, TSVM-RFE exceeded GLAD when unlabelled data along with labelled data were used; it performed much better with gene selection and performed well even if the labelled dataset was small.

On the other hand, TSVM still had some drawbacks when increasing the size of the labelled dataset, as the performance did not significantly improve accordingly. Moreover, when the size of the unlabelled samples was extremely small, the time complexity was correspondingly high.

As with almost all semi-supervised learning algorithms, TSVM showed some instability, as some results of different runs were not the same. This occurred because unlabelled samples may have been wrongly labelled during the learning process. If we find a way in future to select and eliminate the unlabelled sample first, we can then limit the number of newly-labelled samples for re-training the classifiers.